\documentclass[a4paper]{article}

\usepackage{INTERSPEECH2021}

\newcommand\blfootnote[1]{%
  \begingroup
  \renewcommand\thefootnote{}\footnote{#1}%
  \addtocounter{footnote}{-1}%
  \endgroup
}

\title{Improving callsign recognition with air-surveillance data in air-traffic communication}
\name{Iuliia Nigmatulina$^{1,2}$, Rudolf A. Braun$^{1}$, Juan Zuluaga-Gomez$^{1,3}$, Petr Motlicek$^1$}
\address{
  $^1$Idiap Research Institute, Martigny, Switzerland \\
  $^2$Institute of Computational Linguistics, University of Zürich \\
  $^3$Ecole Polytechnique Federale de Lausanne (EPFL), Switzerland
}
\email{\{iuliia.nigmatulina,rudolf.braun,juan-pablo.zuluaga,petr.motlicek\}@idiap.ch}

\begin{document}

\maketitle

\begin{abstract}
Automatic Speech Recognition (ASR) can be used as the assistance of speech communication between pilots and air-traffic controllers. Its application can significantly reduce the complexity of the task and increase the reliability of transmitted information. Evidently, high accuracy predictions are needed to minimise the risk of errors. Especially, high accuracy is required in recognition of key information, such as commands and callsigns, used to navigate pilots. Our results prove that the surveillance data containing callsigns can help to considerably improve the recognition of a callsign in an utterance when the weights of probable callsign n-grams are reduced per utterance. In this paper, we investigate two approaches: (1)~G-boosting, when callsigns weights are adjusted at language model level (G) and followed by the dynamic decoder with an on-the-fly composition, and (2)~lattice rescoring when callsign information is introduced on top of lattices generated using a conventional decoder. Boosting callsign n-grams with the combination of two methods allowed us to gain 28.4\% of absolute improvement in callsign recognition accuracy and up to 74.2\% of relative improvement in WER of callsign recognition.

\end{abstract}

\section{Introduction}
\blfootnote{The work was supported by the European Union’s Horizon 2020 project No. 864702 - ATCO2 (Automatic collection and processing of voice data from air-traffic communications), which is a part of Clean Sky Joint Undertaking. The work was also partially supported by SESAR Joint Undertaking under Grant Agreement No. 884287, under European Union’s Horizon 2020 Research and Innovation programme.}

Communication between pilots and Air-Traffic Controllers (ATCo) is mostly voice-based in order to reduce possible distraction of pilots during the flight and to accelerate information exchange. The voice-based approach, however, makes communication potentially more error-prone because of the amount and speed of information transmission, environmental noise and variability of accents due to the international setting. These factors make the task extremely stressful, pilots and ATCos must always be highly concentrated to guarantee a high level of accuracy to provide safety.

Technical support of pilot-controller communication can increase the level of confidence of correct command perception and reduce the workload. Recent progress in Automatic Speech Recognition (ASR) including online recognition of continuous speech allows new perspectives in improving communication methods between pilots and ATCos~\cite{geacuar2010reducing}.

A possible direction to improve the recognition quality is using contextual information, e.g. surveillance data. One of the key parts of ATCo commands is a callsign \textemdash{} a unique identifier for aircraft, of which the first part is an abbreviation of airline name and the last part is a flight number that contains a digit combination and may also incorporate an additional character combination, e.g. \textit{TVS84J}. The callsign data comes from the radar in a compressed form (first column in Table~\ref{tab:callsigns}), so before using the data for ASR, all callsigns are to be expanded to word sequences (second column in Table~\ref{tab:callsigns}). The compressed form often allows more than one possible realisation in the ATCos' speech: For example, \textbf{DLH5KX} can be expanded as \textit{`hansa five kilo x-ray'} or \textit{`lufthansa five kilo x-ray'}. Since we can not say which particular expansion is true for an utterance callsign, it is important to take all expansion variants into account.

At a certain time point, only few aircraft are usually in the radar zone which means only a limited number of callsigns can be referred to in the ATCo communications. If a recognised callsign does not match any callsign registered by radar at the same time point, it means that there is no corresponding aircraft in the air space and the recognised command is invalid. In the proposed methods, we dynamically introduce surveillance information to increase the probability of recognising those callsigns which are present in the air space at the moment of utterance. In general terms, this research is centered on integrating contextual knowledge during ASR decoding. Nevertheless, previous research has shown that this technique could be also integrated during semi-supervised learning i. e., contextual semi-supervised learning~\cite{zuluaga2021contextual}. The rest of the paper is organised as follows: first, we overview previous research done on the n-gram boosting; then, we give some theoretical background and describe two investigated approaches in more details; then, we present the data and the experiment set up; finally, we report the results and summarise our observations and ideas in the discussion.

\begin{table}[t]
  \caption{Examples of callsigns}
  \label{tab:callsigns}
  \centering
  \begin{tabular}{ ll }
    \toprule
    Callsign & Extended callsign \\
    \midrule
    SWR2689 & swiss two six eight nine \\
    RYR1RK & ryanair one romeo kilo \\
    RYR1SG & ryanair one sierra golf \\
    \bottomrule
  \end{tabular}

\end{table}

\section{Contextual information for improving n-gram recognition}

As a callsign is a sequence of words, using contextual information to improve recognition of callsigns is a task of boosting certain n-grams. When information about currently registered callsigns is available from the surveillance data, it can be used to increase the probability of those particular callsign n-grams in the recognition.

Using contextual information for ATCo ASR has been already used in some previous studies~\cite{shore2012knowledge,oualil2015real}. In~\cite{oualil2015real}, the weighted Levenstein distance is applied as post-processing to overcome the problem of variability of ATCO commands. In~\cite{shore2012knowledge}, a grammar with all semantic concepts of ATC embedded in XML annotation tags is used. After decoding, lattice hypotheses are rescored by adding an additional knowledge source component to the cost function. The knowledge-based rescoring penalises hypotheses which are invalid in the context, e.g. callsigns not registered in the air space. Although this approach helps to considerably increase the recognition accuracy, its limitation is that it deals only with concepts and callsigns which are annotated and included into the grammar. Those n-grams that do not appear in the grammar can not be extracted and evaluated.

In this paper, we propose boosting callsign n-grams by dynamically modifying their weights and adding them, if necessary, in the weighted finite-state transducer (WFST). Similar FST-based approaches with adjusted weights have been applied in different scenarios, for example, improving recognition of named entities (NEs) such as contacts, locations, films, etc. in order to adapt the recognition of voice-driven assistants to users~\cite{hall2015composition,aleksic2015bringing,serrino2019contextual}. In~\cite{aleksic2015bringing}, the authors use class-based Language Model (LM) and bias towards the whole class in a particular context. In~\cite{serrino2019contextual}, potential named entities are firstly detected and tagged with a class, then, the input word lattice is composed with `tagging' FSTs generated for each utterance with a tagged entity. At the final step, contextually relevant NEs are suggested for given phoneme hypotheses inside the time interval corresponding to a tagged path. Approaches in both papers aim to boost a certain pre-defined semantic class and then to take the best recognition candidate within this class. In the situation with callsigns, however, we do not have different pre-defined classes; a list of callsigns to be boosted varies between utterances and should be updated every time.

The approach proposed in~\cite{hall2015composition}, is probably the closest to our lattice rescoring method: a generalised composition is applied to combine baseline LM and n-gram WFSTs generated with the most frequent n-grams from recent news. The main difference is that they perform composition of two FSTs not at the rescoring step but in the first pass decoding as well. As this method achieved good results in recognition of recently frequent in news n-grams, we decided to adapt it for detecting callsign n-grams. In addition to the rescoring method, we also compare the proposed method to the other one involving model modifications.

We address the question of how to boost callsigns by investigating two approaches: 1)~grammar modification with boosting n-grams in G.fst and dynamic graph composition, and 2)~adjusting n-grams weights with lattice rescoring. The former method involves adaptation of the decoding graph itself, when in the latter approach, modifications are applied to the ASR outputs. We compare two methods to each other, as well as to the combination of both. We show that boosting several probable callsigns improves the recognition of a true one. To learn how well the methods work relatively to the ideal scenario, when the correct callsign is known, results of boosting surveillance callsigns are also compared to those achieved with boosting only the ground truth callsign, in our case available from the manual transcriptions.

\section{FST composition}
In a hybrid ASR system the different knowledge sources are represented as WFSTs, which are composed together in the final decoding graph~\cite{mohri2002weighted}:
\begin{equation}
  HCLG = H \circ C \circ L \circ G
  \label{fst-composition}
\end{equation}
where $H, C, L$ and $G$ represent the Hidden Markov Models, context dependency, lexicon and LM (AKA grammar) correspondingly.

The operation of composition maps sequences from input transducers in a way that an output from one transducer should be an input to another one, which allows combining different levels of representation. In other words, with composition one can integrate into a system information from additional knowledge sources.

The complete $HCLG$ graph, however, is very large and complicated and its modification, ideally in real-time, is difficult. Modifying the $G$ is an easier and more flexible approach. The G-boosting method is possible because of new techniques that allow for on-the-fly composition of the $HCL$ and a modified $G$~\cite{novak2012dynamic,hori2007efficient}. For all operations on WFSTs we use the wrapper released as part of \cite{braun2021comparison} in GitHub\footnote{https://github.com/idiap/icassp-oov-recognition}.

\section{Methods}
\subsection{Boosting language model}
In the G-boosting approach, callsign n-grams from the surveillance data are added into the grammar (G.fst) before the decoding step. For each test utterance, the baseline G.fst is modified so that the weights (which are actually costs) of arcs which match the words in a callsign are reduced by a constant $-\log{p}$, or arcs with small weights are created if they do not exist~\cite{braun2021comparison}. As the result, each utterance has its own adjusted G.fst which is then composed on-the-fly with $HCL$ in a lookahead decoding.

\subsection{Lattice rescoring}
\begin{figure}[t]
  \centering
  \includegraphics[width=\linewidth]{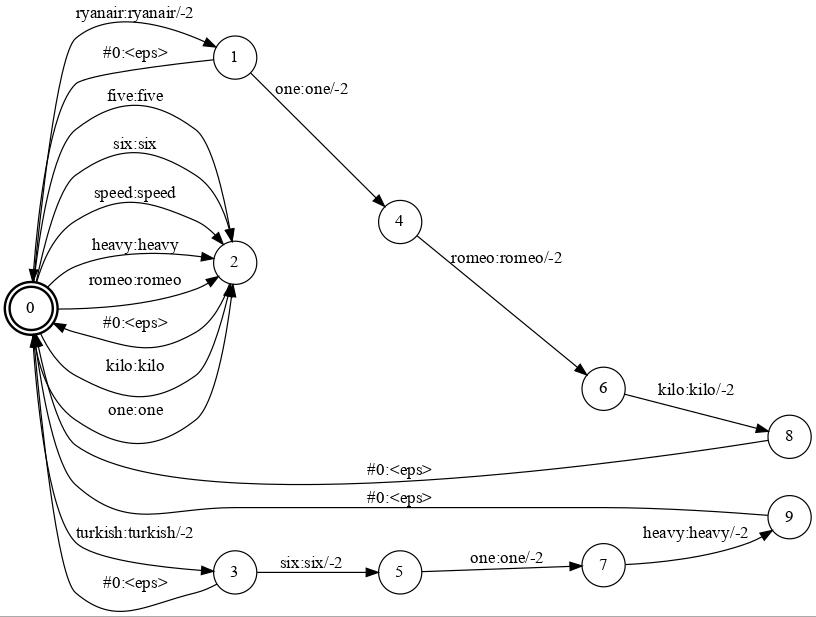}
  \caption{A toy-example of biased FST with callsigns: \textbf{`ryanair one romeo kilo'} and \textbf{`turkish six one heavy'}.}
  \label{fig:pic_fst}
\end{figure}

The lattice rescoring approach is a type of two pass decoding. In first pass, lower order (baseline) LM generates preliminary predictions in the lattice form; in second pass called `rescoring', higher order LM adapts initial hypotheses according to context information. In our case, callsigns n-grams get boosted in the decoding lattices and become more probable to appear in the final predictions. Surveillance information is added after lattices are created at the decoding step. Weights in the lattices are rescored according to the surveillance data: for each test utterance, an FST biased to callsigns n-grams registered at the same timestamp as an utterance is created (see a `toy' example of such \textit{biasing} FST at Fig~\ref{fig:pic_fst}). Then, baseline lattices created in the first pass are converted to FST and composed per each utterance with a corresponding biased callsigns FST. Weights updated in the composition are used for final predictions.

\section{Data and experimental setup}

\subsection{Data}
The ATCo communication is a specific domain, for which only a limited data is available. The collection of ATC data is a challenging task starting from the recording which is often done in a noisy environment and can allow only a medium quality data with frequencies filtered over 4kHz. Additionally, data labelling usually requires highly trained transcribers, mainly with ATCo experience.

Recently, the European CleanSky EC-H2020 ATCO2 project has developed an ASR-based platform for the collection and automatic pre-processing of ATC speech, radar, and surveillance data~\cite{zuluaga2020asratc,zuluaga2020callsign}. The data collected by the project was used for the experiments in this paper\footnote{The ATCO2 data is not publicly available at this moment, as it is considered as personal data. Nevertheless, ATCO2 project aims to release a dataset for research purposes soon.}. 195~hours were used for training the ASR acoustic model (see Section~\ref{subsec:model}) including ATC speech data of different quality and English accents taken from various data sets~\cite{N4NATO,HIWIRE,ATCOSIM,AIRBUS,LDC_ATCC}.

For the callsign boosting experiments, we use four test sets built from the ATCO2 test data with the available surveillance information (Table~\ref{tab:test_sets}). 

\subsubsection{LiveATC data}
The first two test sets are from the LiveATC\footnote{LiveATC.net is primarily a streaming audio network consisting of local receivers tuned to aircraft communications around the world: https://www.liveatc.net/} data recorded from publicly accessible VHF radio channels, which includes both pilots and ATCo speech and, therefore, is of rather low quality (i.e. low SNR often below 10dB)~\cite{zuluaga2020callsign}. One test set is designed in a way that it includes only those utterances that according to the manual transcriptions contain a callsign (\textit{`LiveATC'}). Although such a set is good to test the proposed methods, it does not reflect well the reality when only utterances with callsigns are part of the data. Thus, in addition we employ another test set from the same LiveATC data that includes both types of utterances with and without callsigns (\textit{`LiveATC\_mix'}). 

\subsubsection{MALORCA data}
The other two test sets are prepared with the good quality (i.e. telephone quality speech with SNR usually above 20dB)  data from the MALORCA project\footnote{The Horizon 2020 SESAR project MALORCA (Machine Learning of Speech Recognition Models for Controller Assistance) is partly funded by SESAR Joint Undertaking (Grant Number 698824): https://www.malorca-project.de/wp/.} which includes only ATCo speech. The recognition accuracies of the baseline model are already high above the accuracies reached on VHF LiveATC data (see Table~\ref{tab:results}). The data was collected from the Prague and Vienna airports and, thus, forms two separate sets correspondingly. From the `standard' MALORCA test sets~\cite{srinivasamurthy2017semi} only utterances with the available surveillance information are selected. These utterances are both with and without callsigns.

\begin{table}[t]
  \caption{Test sets with surveillance information (callsigns per utterance \textemdash{} median of callsigns per utterance in the surveillance data)}
  \label{tab:test_sets}
  \centering
  \begin{tabular}{ lcccc }
    \toprule
    & \multicolumn{2}{c}{Num of utterances} & callsigns & \\
    Test set & with & without & per & minutes \\
    & \multicolumn{2}{c}{a callsign} & utterance & \\
    \midrule
    LiveATC & 472 & 0 & 29 & 32 \\
    LiveATC\_mix & 581 & 29 & 28 & 40 \\
    Malorca Prague & 784 & 88 & 5 & 82 \\
    Malorca Vienna & 877 & 38 & 19 & 65 \\
    \bottomrule
  \end{tabular}
\end{table}

%\subsection{Live ATC data inkl. new data from VHF\_RECEIVERS}

%\section{Experimental setup}

\subsection{ASR model}
\label{subsec:model}
For training the baseline acoustic model, as well as for the decoding and rescoring experiments, we used the Kaldi framework~\cite{povey2011kaldi}. The system follows the standard Kaldi recipe, which uses MFCC and i-vectors features. The standard chain training is based on Lattice-free MMI (LF-MMI~\cite{povey2016purely}, which includes 3-fold speed perturbation and one third frame sub-sampling.

The acoustic model is a CNN-TDNNF trained on approximately 1200 hours of ATC labelled data. First, the training databases (195~hours) were augmented by adding noises that match LiveATC audio channel. Afterwards, we applied speed perturbation, obtaining almost 1200 hours of training data. The model was further improved with 700~hours of semi-supervised data collected in LiveATC for different airports from Europe. We follow a standard recipe for semi-supervised training~\cite{Khonglah_ICASSP2020_2020} in Kaldi, which uses a seed model (trained on only the supervised data) to obtain the lattices and best paths of the unsupervised (i.e. 700h LiveATC database). The supervised and unsupervised lattices are then combined to create the new training archives. The LM is 3-gram trained on the same data as the acoustic model with an additional text data coming from additional public resources such as airlines names, airports, ICAO alphabet and way-points in Europe. 

%The acoustic model we used is a CNN-TDNNF model trained on \textcolor{red}{195} hours of the ATC labelled data from different data sets with ivectors and data perturbation, similar to the one described in~\cite{zuluaga2020asratc}. The model was further improved with \textcolor{red}{700} hours of semi-supervised data collected from LiveATC. The LM is 3-gram trained on the same data as the acoustic model with a vocabulary of 28410 words.

\subsection{Evaluation metrics}
For speech recognition for ATCo communication, one of goals is to accurately recognize issued commands by ATCos, where callsigns are one of the key parts. Since this paper focuses on  improving callsign detection, we evaluate proposed methods in two ways: using standard Word Error Rate (WER) for the entire spoken utterances and with WER applied only on the callsign sequences. For the evaluation we use the tool \texttt{texterrors}~\cite{braun2021comparison} where the list of ground truth callsigns per utterance can be passed as an argument. Besides callsign WER, we additionally use accuracy of callsign recognition defined by the percentage of fully correctly recognised callsigns.

\section{Results}

%\textbf{(Local discussion)}
In Table~\ref{tab:results}, the results of the experiments are presented with utterance WER, callsign WER and accuracy of callsign recognition. Combination of both methods achieves the best callsign recognition results for all test sets. G-boosting and lattice rescoring together help to gain up to 4.6\% of absolute (i.e. from 28.0 to 23.4\%), or 16.5\% of relative improvement of utterance WER, as well as up to 28.4\% of absolute improvement in the accuracy of callsigns recognition (see Table~\ref{tab:results}). Table~\ref{tab:improve_examples} gives some examples of improvement where airlines names and callsign word sequences are detected correctly comparing to the baseline predictions.

\begin{table}[t]
  \caption{Results of the boosting experiments (WER \textemdash{} word error rate; CallWER \textemdash{} Callsign WER; Acc \textemdash{} accuracy of callsign recognition)}
  \label{tab:results}
  \centering
  \begin{tabular}{llll}
    \toprule
    \textbf{Model} & \textbf{WER} & \textbf{CallWER} & \textbf{Acc} \\
    \midrule
    \textbf{LiveATC} &  &  & \\
    baseline & 28.0 & 28.5 & 41.3 \\
    rescoring surveillance data & 24.9 & 19.3 & 63.4 \\
    \textit{rescoring ground truth} & 24.7 & 18.0 & 67.6 \\
    G boosting (k=2) & 24.8 & 18.5 & 63.4 \\
    G+rescoring surveillance & \textbf{23.4} & \textbf{14.3} & \textbf{69.7} \\
    \textit{G+rescoring ground truth} & 22.8 & 12.7 & 74.6 \\
    \midrule
    \textbf{LiveATC\_mix} &  &  & \\
    baseline & 30.7 & 29.2 & 50.5 \\
    rescoring surveillance data & 29.5 & 23.9 & 60.8 \\
    \textit{rescoring ground truth} & 27.8 & 17.8 & 72.8 \\
    G boosting (k=4) & 28.1 & 19.5 & 66.2 \\
    G+rescoring surveillance & \textbf{27.2} & \textbf{16.0} & \textbf{71.3} \\
    \textit{G+rescoring ground truth} & 26.3 & 12.2 & 79.8 \\
    \midrule
    \textbf{Malorca Prague} &  &  & \\
    baseline & 3.1 & 2.2 & 94.2 \\
    rescoring surveillance data & \textbf{3.0} & \textbf{1.0} & 97.3 \\
    \textit{rescoring ground truth} & 2.8 & 0.9 & 97.6 \\
    G boosting (k=4) & 3.1 & 1.7 & 95.7 \\
    G+rescoring surveillance & 3.1 & \textbf{1.0} & \textbf{97.4} \\
    \textit{G+rescoring ground truth} & 2.8 & 0.8 & 97.7 \\
    \midrule
    \textbf{Malorca Vienna} &  &  & \\
    baseline & 9.2 & 6.6 & 84.6 \\
    rescoring surveillance data & 8.3 & 3.1 & 93.8 \\
    \textit{rescoring ground truth} & 8.2 & 2.7 & 94.9 \\
    G boosting (k=4) & 8.5 & 3.6 & 91.9 \\
    G+rescoring surveillance & \textbf{8.2} & \textbf{1.7} & \textbf{96.3} \\
    \textit{G+rescoring ground truth} & 8.1 & 1.4 & 97.6 \\
    \bottomrule
  \end{tabular}
\end{table}

As previously discussed, the surveillance data typically provides information about a few callsigns registered in the air space at a certain timestamp. In order to compare our results to the same setup tested with only a {\it ground truth} callsign for each utterance, along with using FSTs biased to all registered callsigns, we repeated the experiments with FSTs biased to a single {\it ground truth} callsign. As information about the only correct callsign is not available in the reality, this is considered as an oracle (upper bound) performance.  Even if the ground truth scores are always noticeably better, systems boosted for all surveillance data callsigns achieve comparable results.

The number of callsigns extracted from radar and used to boost the ASR system per utterance can affect the performance of the methods because by boosting more n-grams it is easier for the system to pick the wrong one in the end. We investigated this effect. The test sets have different numbers of boosted n-grams, from 5 to 29 (see Table~\ref{tab:callsigns}), but even with many boosted callsigns the recognition accuracy goes considerably up comparing to the baseline.

%\textbf{(Noisy vs not-noisy data)}
Our methods also demonstrate consistent results for data of different quality. The level of noise in the recordings of LiveATC and Malorca test sets are very different, as well as the WERs achieved by their baseline systems. Nevertheless, we see considerable improvement for all test sets and the general tendency stays the same.

In addition to lattice rescoring similar to the one described in~\cite{hall2015composition}, we also compare it to adjusting weights inside of a grammar and composing it after with the rest of the graph in a dynamic mode. This method is convenient because it does not require an additional step of rescoring and the generated hypotheses already include biased n-grams. The main inconvenience of G-boosting approach is increasing memory consumption. One has to create a new modified G.fst every time when new information is available. At the same time, the old G.fst is usually kept for the next updates. That doubles the memory usage.

N-gram adaptation language modelling typically involves interpolation of in-domain and out-domain LMs. The strong side of the proposed methods is their flexibility. They adapt current contextual information to improve the recognition of target n-grams without any additional in-domain LMs. This allows using new contextual data per each particular utterance.

\begin{table}[t]
  \caption{Examples of improved callsign recognition (red \textemdash{} wrong; blue \textemdash{} correct)}
  \label{tab:improve_examples}
  \centering
  \begin{tabular}{ ll }
    \toprule
    System & Callsign \\
    \midrule
    Baseline & \textcolor{red}{hello sovar} one nine lima \\
    Boosted & \textcolor{blue}{stobart two} one nine lima \\
    \midrule
    Baseline & ryanair four \textcolor{red}{bye bye} \\
    Boosted & ryanair four \textcolor{blue}{tango mike} \\
    \midrule
    Baseline & \textcolor{red}{one} six \textcolor{red}{zero} three five \\
    Boosted & \textcolor{blue}{airfrans} six \textcolor{blue}{seven} three five \\
    \bottomrule
  \end{tabular}
\end{table} 

\section{Conclusion}

We investigated two methods of integrating contextual radar data in order to dynamically improve the recognition of callsigns per utterance. The first approach involves modifying weights for callsign n-grams in the grammar component of the decoding graph ($G$) followed by the dynamic lookahead decoding. In the second approach, callsign n-grams are boosted by composing the decoded lattices (lattices generated in the first pass decoding and converted to WFSTs) and contextual WFSTs biased to the callsign sequences registered in the surveillance data.

The best results are achieved with the combination of both methods. The same tendency of improvement is noticeable in all test sets and in the recordings of both lower and higher quality with the relative improvement of callsign recognition varying from 45.2 to 74.2\% depending on the test set.

Introduction of contextual information considerably improves recognition of callsigns and, therefore, recognition of ATCo commands. As a noisy environment leading to lower recognition accuracy is often a reality in pilot-ATCo communication, the proposed methods and their combination will definitely improve the recognition of the key information in ATCo speech. 

%\section{Acknowledgements}
% It is in the first page now (i added Petr comments)

\bibliographystyle{IEEEtran}

\bibliography{references}

\begin{thebibliography}{10}
\providecommand{\url}[1]{#1}
\csname url@samestyle\endcsname
\providecommand{\newblock}{\relax}
\providecommand{\bibinfo}[2]{#2}
\providecommand{\BIBentrySTDinterwordspacing}{\spaceskip=0pt\relax}
\providecommand{\BIBentryALTinterwordstretchfactor}{4}
\providecommand{\BIBentryALTinterwordspacing}{\spaceskip=\fontdimen2\font plus
\BIBentryALTinterwordstretchfactor\fontdimen3\font minus
  \fontdimen4\font\relax}
\providecommand{\BIBforeignlanguage}[2]{{%
\expandafter\ifx\csname l@#1\endcsname\relax
\typeout{** WARNING: IEEEtran.bst: No hyphenation pattern has been}%
\typeout{** loaded for the language `#1'. Using the pattern for}%
\typeout{** the default language instead.}%
\else
\language=\csname l@#1\endcsname
\fi
#2}}
\providecommand{\BIBdecl}{\relax}
\BIBdecl

\bibitem{geacuar2010reducing}
C.-M. Geac{\u{a}}r, ``Reducing pilot/atc communication errors using voice
  recognition,'' in \emph{Proceedings of ICAS}, 2010.

\bibitem{zuluaga2021contextual}
\BIBentryALTinterwordspacing
J.~Zuluaga-Gomez, I.~Nigmatulina, A.~Prasad, P.~Motlicek, K.~Vesel{\`y},
  M.~Kocour, and I.~Sz{\"o}ke, ``Contextual semi-supervised learning: An
  approach to leverage air-surveillance and untranscribed atc data in asr
  systems,'' in \emph{Interspeech 2021, 22st Annual Conference of the
  International Speech Communication Association, Virtual Event, Brno,
  Czechia}.\hskip 1em plus 0.5em minus 0.4em\relax {ISCA}, 2021. [Online].
  Available: \url{https://arxiv.org/abs/2104.03643}
\BIBentrySTDinterwordspacing

\bibitem{shore2012knowledge}
T.~Shore, F.~Faubel, H.~Helmke, and D.~Klakow, ``Knowledge-based word lattice
  rescoring in a dynamic context,'' in \emph{Thirteenth Annual Conference of
  the International Speech Communication Association}, 2012.

\bibitem{oualil2015real}
Y.~Oualil, M.~Schulder, H.~Helmke, A.~Schmidt, and D.~Klakow, ``Real-time
  integration of dynamic context information for improving automatic speech
  recognition,'' in \emph{Sixteenth Annual Conference of the International
  Speech Communication Association}, 2015.

\bibitem{hall2015composition}
K.~Hall, E.~Cho, C.~Allauzen, F.~Beaufays, N.~Coccaro, K.~Nakajima, M.~Riley,
  B.~Roark, D.~Rybach, and L.~Zhang, ``Composition-based on-the-fly rescoring
  for salient n-gram biasing,'' 2015.

\bibitem{aleksic2015bringing}
P.~Aleksic, M.~Ghodsi, A.~Michaely, C.~Allauzen, K.~Hall, B.~Roark, D.~Rybach,
  and P.~Moreno, ``Bringing contextual information to google speech
  recognition,'' 2015.

\bibitem{serrino2019contextual}
J.~Serrino, L.~Velikovich, P.~S. Aleksic, and C.~Allauzen, ``Contextual
  recovery of out-of-lattice named entities in automatic speech recognition.''
  in \emph{Interspeech}, 2019, pp. 3830--3834.

\bibitem{mohri2002weighted}
M.~Mohri, F.~Pereira, and M.~Riley, ``Weighted finite-state transducers in
  speech recognition,'' \emph{Computer Speech \& Language}, vol.~16, no.~1, pp.
  69--88, 2002.

\bibitem{novak2012dynamic}
J.~R. Novak, N.~Minematsu, and K.~Hirose, ``Dynamic grammars with lookahead
  composition for wfst-based speech recognition,'' in \emph{Thirteenth Annual
  Conference of the International Speech Communication Association}, 2012.

\bibitem{hori2007efficient}
T.~Hori, C.~Hori, Y.~Minami, and A.~Nakamura, ``Efficient wfst-based one-pass
  decoding with on-the-fly hypothesis rescoring in extremely large vocabulary
  continuous speech recognition,'' \emph{IEEE Transactions on audio, speech,
  and language processing}, vol.~15, no.~4, pp. 1352--1365, 2007.

\bibitem{braun2021comparison}
R.~A. Braun, S.~Madikeri, and P.~Motlicek, ``A comparison of methods for
  oov-word recognition on a new public dataset,'' in \emph{ICASSP 2021-2021
  IEEE International Conference on Acoustics, Speech and Signal Processing
  (ICASSP)}.\hskip 1em plus 0.5em minus 0.4em\relax IEEE, 2021, pp. 5979--5983.

\bibitem{zuluaga2020asratc}
\BIBentryALTinterwordspacing
J.~Zuluaga{-}Gomez, P.~Motl{\'{\i}}cek, Q.~Zhan, K.~Vesel{\'{y}}, and R.~Braun,
  ``Automatic speech recognition benchmark for air-traffic communications,'' in
  \emph{Interspeech 2020, 21st Annual Conference of the International Speech
  Communication Association, Virtual Event, Shanghai, China, 25-29 October
  2020}.\hskip 1em plus 0.5em minus 0.4em\relax {ISCA}, 2020, pp. 2297--2301.
  [Online]. Available: \url{https://doi.org/10.21437/Interspeech.2020-2173}
\BIBentrySTDinterwordspacing

\bibitem{zuluaga2020callsign}
J.~Zuluaga-Gomez, K.~Vesel{\`y}, A.~Blatt, P.~Motlicek, D.~Klakow, A.~Tart,
  I.~Sz{\"o}ke, A.~Prasad, S.~Sarfjoo, P.~Kol{\v{c}}{\'a}rek \emph{et~al.},
  ``Automatic call sign detection: Matching air surveillance data with air
  traffic spoken communications,'' in \emph{Multidisciplinary Digital
  Publishing Institute Proceedings}, vol.~59, no.~1, 2020, p.~14.

\bibitem{N4NATO}
S.~Pigeon, W.~Shen, A.~Lawson, and D.~A.~v. Leeuwen, ``Design and
  characterization of the non-native military air traffic communications
  database (nnmatc),'' in \emph{Eighth Annual Conference of the International
  Speech Communication Association}, 2007.

\bibitem{HIWIRE}
J.~Segura, T.~Ehrette, A.~Potamianos, D.~Fohr, I.~Illina, P.~Breton, V.~Clot,
  R.~Gemello, M.~Matassoni, and P.~Maragos, ``The hiwire database, a noisy and
  non-native english speech corpus for cockpit communication,'' \emph{Online.
  http://www. hiwire. org}, 2007.

\bibitem{ATCOSIM}
K.~Hofbauer, S.~Petrik, and H.~Hering, ``The atcosim corpus of non-prompted
  clean air traffic control speech.'' in \emph{LREC}, 2008.

\bibitem{AIRBUS}
E.~Delpech, M.~Laignelet, C.~Pimm, C.~Raynal, M.~Trzos, A.~Arnold, and
  D.~Pronto, ``{A Real-life, French-accented Corpus of Air Traffic Control
  Communications},'' in \emph{Proceedings of the Eleventh International
  Conference on Language Resources and Evaluation (LREC 2018)}, 2018.

\bibitem{LDC_ATCC}
\BIBentryALTinterwordspacing
J.~Godfrey, ``{The Air Traffic Control Corpus (ATC0) - LDC94S14A},'' 1994.
  [Online]. Available: \url{https://catalog.ldc.upenn.edu/LDC94S14A}
\BIBentrySTDinterwordspacing

\bibitem{srinivasamurthy2017semi}
A.~Srinivasamurthy, P.~Motlicek, I.~Himawan, G.~Szaszak, Y.~Oualil, and
  H.~Helmke, ``Semi-supervised learning with semantic knowledge extraction for
  improved speech recognition in air traffic control,'' in \emph{Proc. of the
  18th Annual Conference of the International Speech Communication
  Association}, 2017.

\bibitem{povey2011kaldi}
D.~Povey, A.~Ghoshal, G.~Boulianne, L.~Burget, O.~Glembek, N.~Goel,
  M.~Hannemann, P.~Motlicek, Y.~Qian, P.~Schwarz \emph{et~al.}, ``The kaldi
  speech recognition toolkit,'' in \emph{IEEE workshop on automatic speech
  recognition and understanding}, no. CONF.\hskip 1em plus 0.5em minus
  0.4em\relax IEEE Signal Processing Society, 2011.

\bibitem{povey2016purely}
D.~Povey, V.~Peddinti, D.~Galvez, P.~Ghahremani, V.~Manohar, X.~Na, Y.~Wang,
  and S.~Khudanpur, ``Purely sequence-trained neural networks for asr based on
  lattice-free mmi.'' in \emph{Interspeech}, 2016, pp. 2751--2755.

\bibitem{Khonglah_ICASSP2020_2020}
B.~Khonglah, S.~Madikeri, S.~Dey, H.~Bourlard, P.~Motlicek, and J.~Billa,
  ``Incremental semi-supervised learning for multi-genre speech recognition,''
  in \emph{Proceedings of ICASSP 2020}, 2020.

\end{thebibliography}

% \begin{thebibliography}{9}
% \bibitem[1]{Davis80-COP}
%   S.\ B.\ Davis and P.\ Mermelstein,
%   ``Comparison of parametric representation for monosyllabic word recognition in continuously spoken sentences,''
%   \textit{IEEE Transactions on Acoustics, Speech and Signal Processing}, vol.~28, no.~4, pp.~357--366, 1980.
% \bibitem[2]{Rabiner89-ATO}
%   L.\ R.\ Rabiner,
%   ``A tutorial on hidden Markov models and selected applications in speech recognition,''
%   \textit{Proceedings of the IEEE}, vol.~77, no.~2, pp.~257-286, 1989.
% \end{thebibliography}

\end{document}